\documentclass[11pt]{asaproc}

\usepackage{rotating}
\usepackage{caption}
\usepackage{amsfonts}
\usepackage{lscape}
\usepackage{amsmath}
\usepackage{graphicx,psfrag,epsf}
\usepackage{enumerate}
\usepackage{url} 
\usepackage{mathrsfs}
\usepackage{amssymb}
\usepackage{float}
\usepackage{setspace}
\usepackage{booktabs}
\usepackage{url}
\usepackage{color}
\usepackage{amsmath}
\usepackage{bm}
\usepackage[numbers,sort&compress]{natbib}
\usepackage{pdfpages}
\usepackage{multirow}
\graphicspath{{figures/}}
\usepackage{amssymb}
\usepackage{rotating}

\usepackage{fancyhdr}

\usepackage{times}

\title{ Regression-Enhanced Random Forests}

\author{
	Haozhe Zhang\thanks{Department of Statistics, Iowa State University, Ames, IA 50011} \and
	 Dan Nettleton\thanks{Department of Statistics, Iowa State University, Ames, IA 50011} \and Zhengyuan Zhu\thanks{Department of Statistics, Iowa State University, Ames, IA 50011}}
\begin{document}

\maketitle
\thispagestyle{fancy}

\begin{abstract}
Random forest (RF) methodology is one of the most popular machine learning techniques for prediction problems. In this article, we discuss some cases where random forests may suffer and propose a novel generalized RF method, namely regression-enhanced random forests (RERFs), that can improve on RFs by borrowing the strength of penalized parametric regression. The algorithm for constructing RERFs and selecting its tuning parameters is described. Both simulation study and real data examples show that RERFs have better predictive performance than RFs in important situations often encountered in practice. Moreover, RERFs may incorporate known relationships between the response and the predictors, and may give reliable predictions in extrapolation problems where predictions are required at points out of the domain of the training dataset. Strategies analogous to those described here can be used to improve other machine learning methods via combination with penalized parametric regression techniques.
\end{abstract}

\section{Introduction}
Random forest (RF) methodology, proposed by L. Breiman \cite{breiman2001random}, is one of the most popular machine learning techniques for regression and classification problems. In the last few years, there have been many methodological and theoretical advances in the random forests approach. Some methodological developments and extensions include case-specific random forests \cite{xu2016case}, multivariate random forests \cite{segal2011multivariate}, quantile regression forests \cite{meinshausen2006quantile}, random survival forests \cite{ishwaran2008random}, enriched random forests for microarry data \cite{amaratunga2008enriched} and predictor augmentation in random forests \cite{xu2014predictor} among others. For theoretical developments, the statistical and asymptotic properties of random forests have been intensively investigated. Advances have been made in the areas such as consistency \cite{biau2008consistency} \cite{ scornet2015consistency}, variable selection \cite{genuer2010variable} and the construction of confidence intervals \cite{wager2014confidence}.

Although RF methodology has proven itself to be a reliable predictive approach in many application areas \cite{biau2016random}\cite{he2016data}, there are some cases where random forests may suffer. First, as a fully nonparametric predictive algorithm, random forests may not efficiently incorporate known relationships between the response and the predictors. Second, random forests may fail in extrapolation problems where predictions are required at points out of the domain of the training dataset. For regression problems, a random forest prediction is an average of the predictions produced by the trees in the forest. Because each tree prediction corresponds to some weighted average of the response values $Y_{1},\ldots, Y_{n}$ observed in the original training data, we can view the final random forest prediction at some observed predictor vector $\bm{X}_{0}$ as a convex combination of the training response values given by
\begin{equation}
\widehat{Y}(\bm{X}_{0})=\sum_{i=1}^{n}w_{i}(\bm{X}_{0})Y_{i},
\end{equation}
where $w_{i}(\bm{X}_{0}), \ldots, w_{n}(\bm{X}_{0})$ are nonnegative weights with the constraint $\sum_{i=1}^{n}w_{i}(\bm{X}_{0})=1$. It follows that 
\begin{equation} \label{bounds}
\underset{1\leq i \leq n}{\mathrm{min}}Y_{i} \leq \widehat{Y}(\bm{X}_{0}) \leq \underset{1\leq i \leq n}{\mathrm{max}}Y_{i}.
\end{equation}
As a consequence, the predictions given by random forests are always within the range of response values in the training dataset, which is problematic if the response values in the target dataset tend to fall outside this range.
%

We illustrate the above issues by considering the problem of forecasting Iowa corn yield. The dataset, that will be further introduced in Section \ref{example2}, contains county-level corn yield data and predictor variables that provide information about soil quality and environmental conditions during 28 growing seasons. We used random forests to forecast corn yields in the coming year by using the yield and predictor data in previous years as a training dataset. Random forests failed to outperform multivariate linear regression in this problem. For example, the root mean square error (RMSE) of random forests for predicting 2015 corn yield was slightly more than $10\%$ higher than the RMSE of multivariate linear regression. 


There are at least two reasons why multivariate linear regression outperforms random forests for predicting Iowa corn yield. First, in some years, the weather was so hot and dry that the values of temperature and precipitation were beyond the ranges of those in the training dataset, which creates an extrapolation problem. Second, corn yield has been increasing generally over time due to consistent genetic improvement of maize and agricultural technology developments. When forecasting corn yield for a future year using random forests,  as shown by (\ref{bounds}), each forecast is bounded above by the largest corn yield in the training dataset, even if the past trend suggests a record-setting crop for that future year.

Next we use a simulated example to illustrate this point. Let the data-generating model be $Y=f(\bm{X})+10Z+\epsilon$, where $Y$ is the response variable, $\bm{X}=(X_{1},\ldots,X_{10})$ and $Z$ are the predictors, and $\epsilon$ is a mean-zero error term. Suppose 
\begin{equation}
f(\bm{X})=0.1e^{4x_{1}}+\frac{4}{1+e^{-20(x_{2}-1/2)}}+3x_{3}+2x_{4}+x_{5}+0\times\sum_{i=6}^{10}x_{i},
\end{equation}
a partially nonlinear additive function that is Equation (56) in J. H. Friedman's ``MARS" paper \cite{friedman1991multivariate}. We want to predict $Y$ by using the predictors $\bm{X}$ and $Z$. The distributions of predictors $\bm{X}$ in both training and validation datasets are identical, and they are independently simulated from the uniform distribution $\text{unif}(0,1)$. In the training dataset, the predictor variable $Z$ is sampled from $\text{unif} (0,0.8)$, while $Z$ is sampled from $\text{unif} (0,1)$ in the validation dataset. The sample sizes for the training and the validation datasets are $1500$ and $300$, respectively.

\begin{figure}[ht]
	\centering
	\includegraphics[scale=0.7]{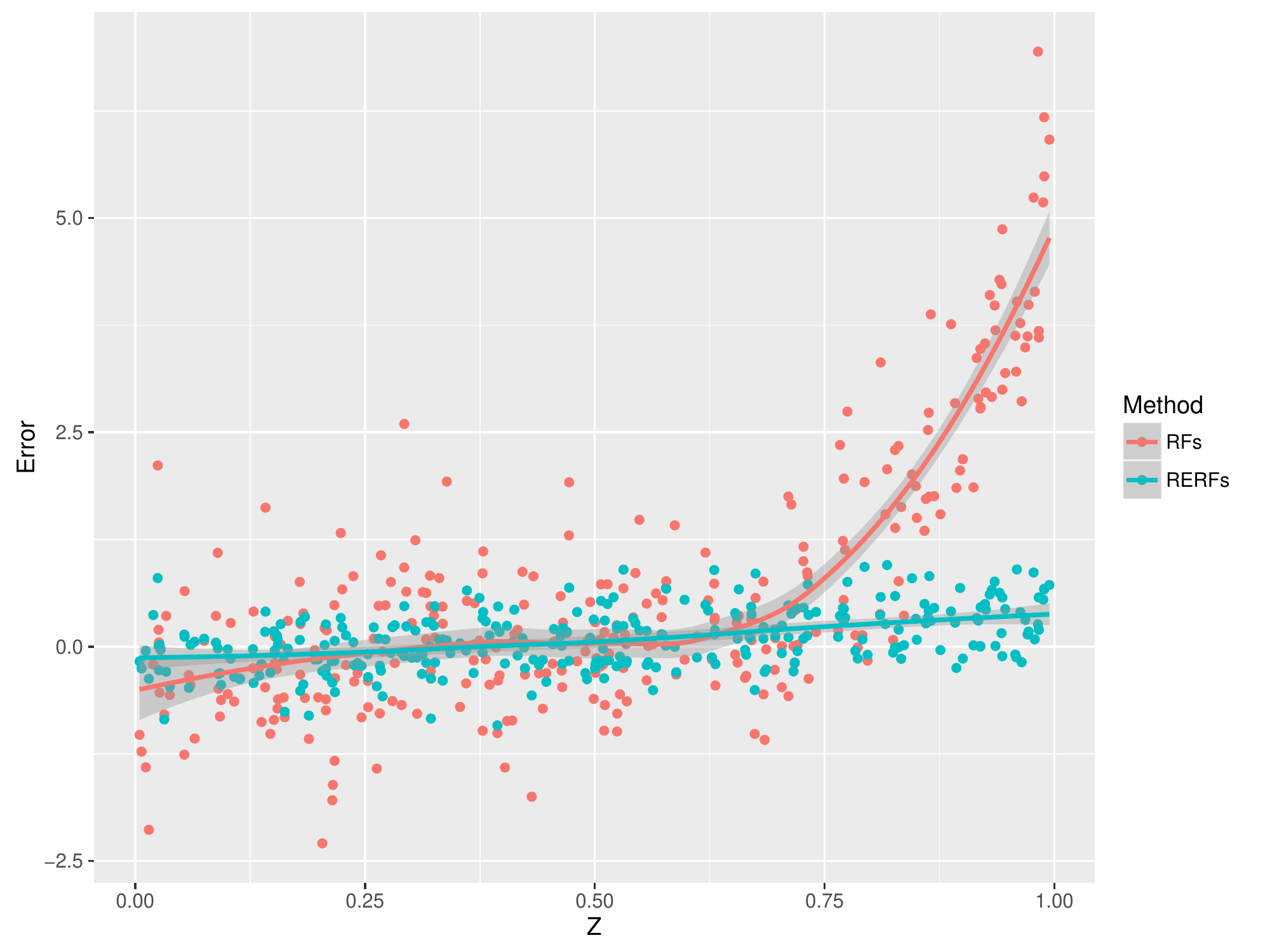}
	\caption{The pointwise prediction errors $Y-\widehat{Y}$ given by a random forest (red) and a regression-enhanced random forest (blue) against the predictor $Z$ and the corresponding Loess smooth curves of $Y-\widehat{Y}$ against $Z$.}\label{introduction_figure}
\end{figure}

Figure \ref{introduction_figure} presents prediction errors when analyzing the simulated data with a random forest and with a regression-enhanced random forest (RERF), the method we introduce in this paper. The red points and the red smoothed curve in the Figure \ref{introduction_figure} illustrate the relationship between the predictor $Z$ and the pointwise prediction errors $Y-\widehat{Y}$ given by the standard RFs. When the predictor $Z > 0.8$, the predicted errors are relatively large. This example indicates that the random forest approach suffers in linear extrapolation.

To address the challenge, we develop RERF methodology that generalizes RF methodology by combining penalized parametric regression with RFs. The purpose of this paper is to introduce RERFs and investigate the prediction performance of RERFs in comparison with RFs. Parametric methods, such as multivariate linear regression and Lasso, can parsimoniously account for scientific mechanisms that dictate approximate linear relationships between the response and predictor variables and allow for effective extrapolation out of the training dataset domain \cite{breiman2001statistical}. Nonparametric machine learning algorithms, such as random forests and artificial neural networks, can account for nonlinearity and complex factor interactions. RERFs can capitalize on the strength of the two types of methods and overcome the corresponding disadvantages. The simulation study and the two real data examples in this paper reach the same conclusion: the prediction performance of RERFs is better than that of the standard RFs in both interpolation and extrapolation problems. Furthermore, in extrapolation cases, RERFs far outperform RFs.

The rest of the paper is organized as follows. Section 2 introduces the algorithm
for building RERFs and also discusses tuning parameter selection. In Section 3, we conduct a simulation study to examine the prediction performance of RERFs in comparison to RFs and Lasso. To illustrate the proposed methodology and demonstrate its relevance in
practice, Section \ref{example} provides two real data examples involving high-performance concrete strength prediction and Iowa corn yield forecasting. We conclude with further discussion of RERF methodology in Section \ref{discussion}.

\section{Method} \label{Method}
Regression-enhanced random forests (RERFs) is a hybrid of random forests and penalized parametric regression. RERFs can improve random forests in prediction accuracy and also incorporate known relationships between the response variable and the predictors. Penalized parametric regression involves a penalty function $P(\cdot)$ applied on the regression coefficient $\bm{\beta}$, which amounts to solving a minimization problem of the form
\begin{equation}
\underset{\bm{\beta}}{\textrm{min}}\{L(Y_{i},\widehat{Y}_{i})+\lambda P(\bm{\beta})\},
\end{equation}
where $L(\cdot, \cdot)$ is a loss function and $\widehat{Y}_{i}$ is the predicted response value that can be regarded as a function of $\beta$. $\lambda$ is called the penalty parameter. Lasso with an $\ell^{1}$ penalty function and Ridge regression with an $\ell^{2}$ penalty function are two examples of penalized parametric regression. Penalized parametric regression can be used to capture the global trend and incorporate scientific knowledge about linear structure, but may not be flexible enough to accommodate nonlinearity.  Random forests offer a flexible nonparametric approach for prediction, which leads to small fitting errors compared with parametric methods. However, small fitting errors do not necessarily imply small prediction errors, especially in extrapolation problems. As shown by Figure \ref{introduction_figure} and the simulated example in Section 1, random forests may suffer in extrapolation problems.

Let $Y$ be a continuous response variable and $\bm{X}$ a $p$-dimensional vector of predictor variables. We assume a standard data-generating model given by 
\begin{equation}
Y = f(\bm{X}) +\epsilon
\end{equation}
for both training and validation datasets. We assume a training dataset $\pmb{C}=\{C_{i}=(\bm{X}_{i}, Y_{i}):i=1,\ldots, N\}$ with a sample size $N$ is available to fit the model for prediction. The random forests algorithm has two tuning parameters \cite{breiman2001random}, mtry and nodesize,  denoted as $m$ and $s$. The RERF algorithm is described as follows:\\


\noindent {\bf Regression-Enhanced Random Forest Algorithm}
\begin{itemize}
	\item Step 1: Extend the $p$-dimensional predictor $\bm{X}$ to a $(p+q)$-dimensional predictor $\bm{X}^{*}$ by adding higher-order, interaction or other known parametric functions of $\bm{X}$.
	\item Step 2: Run Lasso of $Y$ on $\bm{X}^{*}$ with a pre-specified penalty parameter $\lambda$. Let $\bm{\widehat{\beta}}_{\lambda}$ be the estimated coefficient, and $\epsilon^{\lambda}=Y-\bm{X}^{*}\bm{\widehat{\beta}}_{\lambda}$ be the residual from Lasso. Create a new training dataset $\bm{C}^{\lambda}=\{C^{\lambda}_{i}=(\bm{X}_{i},\epsilon^{\lambda}_{i}):i=1,\cdots, N\}$.
	\item Step 3: Build a random forest $\bm{T}_{m,s}$ using $\bm{C}^{\lambda}$ with pre-specified $m$ and $s$. A prediction for the response at a given predictor value $\bm{X}_{0}$ is  $\widehat{Y}(\bm{X}_{0})=\bm{X}_{0}\bm{\widehat{\beta}}_{\lambda}+\bm{T}_{m,s}(\bm{X}_{0})$.
	\item Step 4: Select the tuning parameters ($\lambda, m, s$) through $k$-fold cross validation by repeating step 2 and step 3 for candidate values of $\lambda$, $m$, and $s$. The selected tuning parameters are denoted by $\lambda^{*}, m ^{*}$ and $s^{*}$.
	\item Step 5: The RERF prediction for the response at $\bm{X}_{0}$ is given by  $\widehat{Y}(\bm{X}_{0})=\bm{X}_{0}\bm{\widehat{\beta}}_{\lambda^{*}}+\bm{T}_{m^{*},s^{*}}(\bm{X}_{0})$.
\end{itemize}

To explain the mechanics of RERFs, we will discuss each step in the algorithm in detail. Expanding the design matrix in step 1 is optional. Whether to add higher-order, interaction or other parametric terms should be decided by exploratory analysis or knowledge of the relationship between $Y$ and $\bm{X}$. The main aim of Lasso regression in Step 2 is to select variables in order to find a parametric structure that incorporates the global trend and known relationships between the response and predictors. The penalty parameter $\lambda$ controls the strength of variable selection. When $\lambda=0$, Lasso regression in Step 2 is equivalent to multivariate regression without regularization. When $\lambda \rightarrow \infty$, Lasso regression in Step 2 is equivalent to regressing on a constant value, i.e., fitting an intercept-only model. Thus, RERFs will be reduced to RFs for sufficiently large $\lambda$, and RFs can be viewed as a special case of RERFs.

Tuning parameter selection in Step 4 is critical to the performance of RERFs. As a hybrid method of Lasso and RFs, regression-enhanced random forests have three tuning parameters: Lasso penalty parameter ($\lambda$), nodesize ($s$) and mtry ($m$).  Our approach for simultaneous selection of these three tuning parameters is an exhaustive search on 3-dimensional tuning parameter space. The value of $\lambda$ plays an important role in the prediction performance of RERFs. The optimal value of $\lambda$ for RERFs is not determined based on the cross-validation performance of the Lasso predicton. Instead, the optimal value of $\lambda$ is determined based on the cross-validation performance of the RERF, which involves building a RF with Lasso residuals as response values. The plausible values of $\lambda$ are positive and unbounded. In our numerical examples, we choose $\lambda$ from among the values in the set $\{\exp (\log(0.001)+h\times\frac{\log(100)-\log(0.001)}{99}): h =0,\ldots, 99\}$, which is a set of 100 points from 0 to 99 equally spaced on the logarithm scale. Following the advice of Breiman as recounted by \cite{liaw2002classification}, we consider mtry equal to the default value of $\max\{1, \lfloor p/3\rfloor\}$ as well as half and twice the default value. For nodesize, we consider the default value of $5$ as well as $1$ (the value recommended by Breiman for classification problems). Throughout the paper, all the results from RERFs, Lasso and RFs were obtained by selecting tuning parameters by cross validation.

\section{Simulation study} \label{simulation}
In this section, we conduct a simulation study to examine the prediction performance of RERFs compared with RFs for both interpolation and extrapolation cases. We simulated data from a data-generating model given by 
\begin{equation}
Y = f(\bm{X}) +\epsilon.
\end{equation}
The independent random errors $\epsilon$ follow $N(0,0.5^{2})$. We considered three different structures for $f(\cdot)$:
\begin{itemize}
	\item $L$: a linear model with an additive structure
	\begin{equation}
	f(\bm{X})=x_{1}+x_{2}+2x_{3}+2x_{4}+0\sum_{i=5}^{10}x_{i},
	\end{equation}
	\item $P$: a partially linear model with an additive structure
	\begin{equation}
	f(\bm{X})=\sin(\pi x_{1})+\frac{4}{1+e^{-20x_{2}+10}}+2x_{3}+2x_{4}+0\sum_{i=5}^{10}x_{i},
	\end{equation}
	\item $N$: a non-additive partially linear model
	\begin{equation}
	f(\bm{X})=\sin(\pi x_{1})+\frac{4}{1+e^{-20x_{2}+10}}+2x_{3}+2x_{4}+3x_{3}x_{4}+0\sum_{i=5}^{10}x_{i}.
	\end{equation}
	
\end{itemize}
We also considered two different sampling distributions for $\bm{X}$ that lead to interpolation ($I$) and extrapolation ($E$) as follows:
\begin{itemize}
	\item $I$: all 10 predictor observations are i.i.d.\ $\textrm{unif}(0,1)$ in both training and validation datasets.
	\item $E$: $x_{3}$ observations are i.i.d.\ $\textrm{beta}(4,8)$ in the training dataset and  i.i.d.\ $\textrm{beta}(5,1)$ in the validation dataset, for which predictions of $Y$ are desired. The other 9 predictor observations are i.i.d.\ $\textrm{unif}(0,1)$ in both training and validation datasets.
\end{itemize}
Most of the observations generated from $\textrm{beta}(4,8)$ are less than $0.6$, while most of those generated from $\textrm{beta}(5,1)$ are larger than $0.6$. Thus, prediction for the second case ($E$) often involves extrapolation.

In total, we consider 6 simulation scenarios formed by all the combinations of $f(\cdot)$ and the distributions for $\bm{X}$ labeled as $L\times I$, $P\times I$, $N\times I$, $L\times E$, $P \times E$ and $N\times E$.
For each scenario, 1000 simulation runs were conducted. In each run, 1000 training observations and 100 validation observations were randomly and independently generated from the joint distribution of $(\bm{X},Y)$. We trained Lasso, RFs and RERFs using training data. For each simulated training dataset, values of the tuning parameters for Lasso ($\lambda$), RFs ($s$, $m$) and RERFs ($\lambda$, $s$, $m$) were separately selected using cross-validation. Then we predicted response values for the validation data. Finally, the root mean square errors (RMSEs) were then calculated over the validation dataset. 

A variety of implementations of random forests have been developed. We use the R package \textit{randomForest}  \cite{liaw2002classification} to  implement random forests algorithm throughout this paper. This package was derived from the Fortran code originally by Leo Breiman and Adele Cutler \cite{xu2016case}.  Lasso regression was implemented via the R Package \textit{glmnet} \cite{friedman2010regularization}.

\begin{figure}[ht]
	\centering
	\includegraphics[scale=0.7]{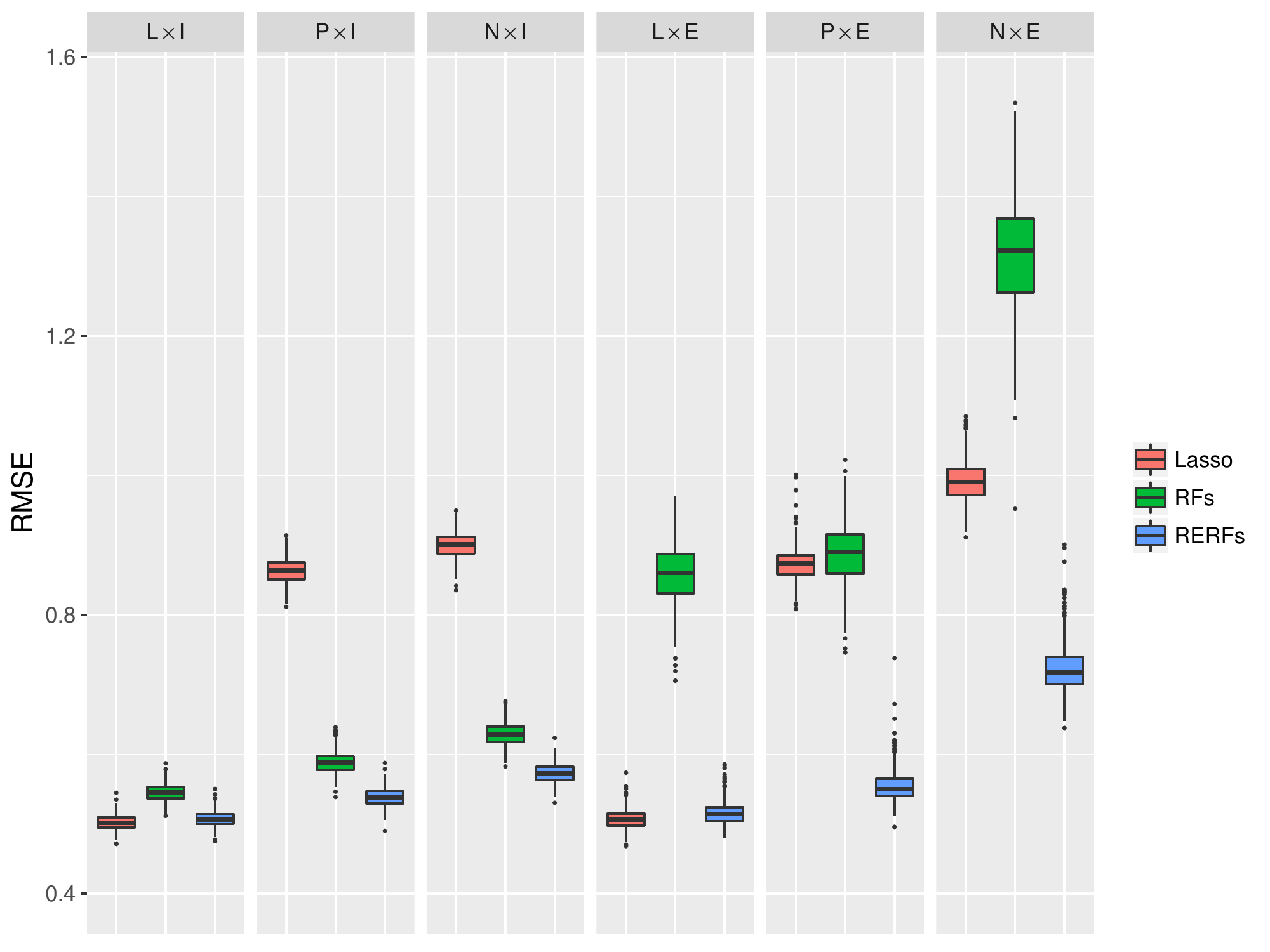}
	\caption{Boxplots of the RMSE values for Lasso, RFs, and RERFs for each data-generating model in the simulation study of Section \ref{simulation}.}\label{simulation_rmse}
\end{figure}

The RMSEs from the simulation are shown in Figure \ref{simulation_rmse}.  RERFs exhibited lower RMSE than RFs for both interpolation and extrapolation problems, regardless of whether Lasso performed better than RFs or not. Particularly in extrapolation cases, RERFs far outperformed RFs. The performance of RERFs was better than that of Lasso for all the models except $L\times I$ and $L\times E$. Because $L\times I$ and $L\times E$ are both linear models, Lasso is expected to perform well for these cases. Nevertheless, the RMSEs of Lasso and RERFs were very close for $L\times I$ and $L\times E$.

\begin{figure}[ht]
	\centering
	\includegraphics[scale=0.7]{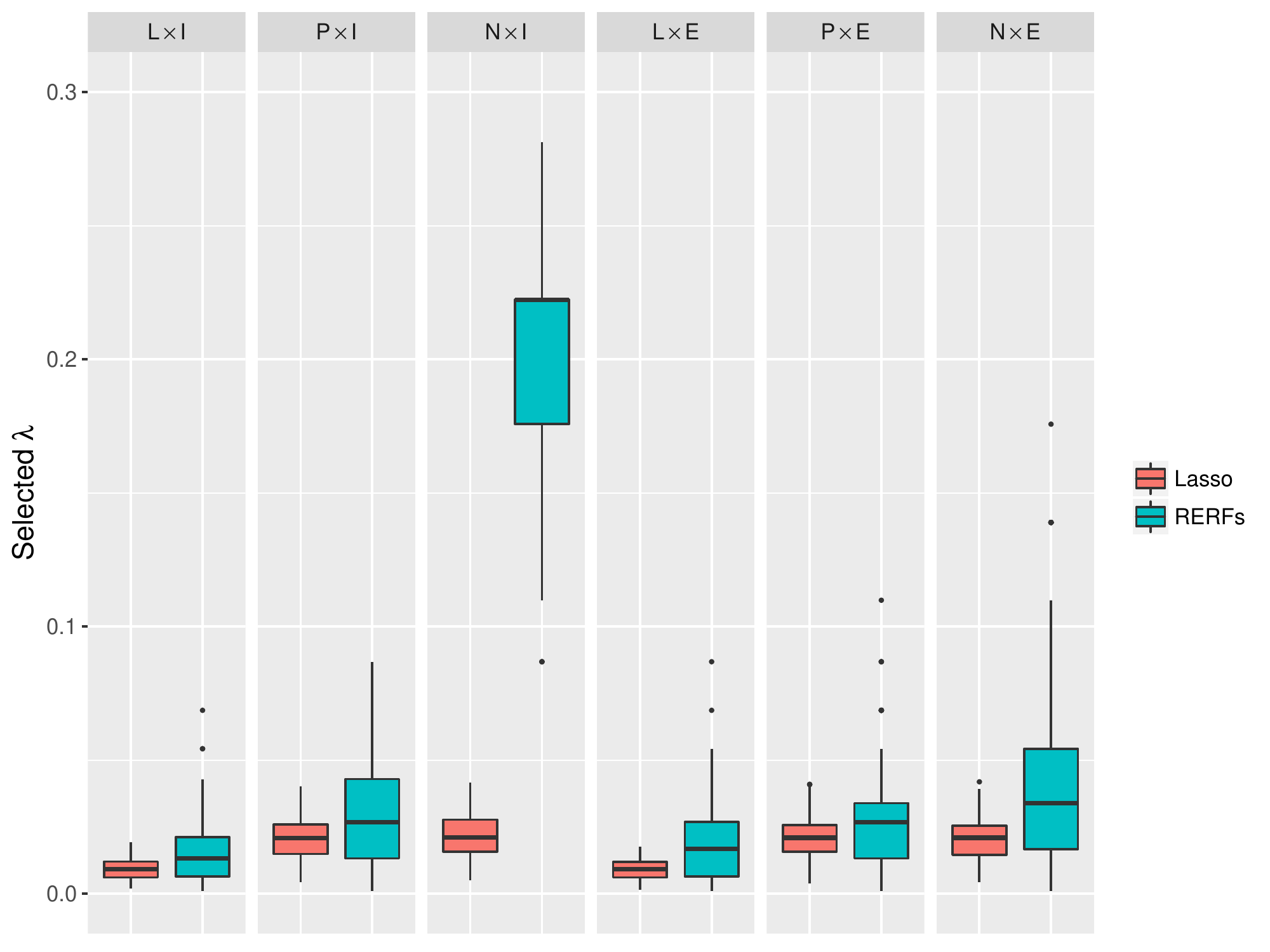}
	\caption{Boxplots of the selected penalty parameter values for Lasso and RERFs for each data-generating model in the simulation study of Section \ref{simulation}.}\label{simulation_lambda}
\end{figure}

As stated in Section \ref{Method}, the value of penalty parameter $\lambda$ has a substantial effect on the prediction performance of RERFs. Figure \ref{simulation_lambda} reports the selected penalty parameter values for Lasso and RERFs for different data-generating models. In general, the selected penalty parameter values of RERFs are larger than those of Lasso. Picking one simulated dataset from Scenario $N\times I$ as an example, the selected penalty parameter value is $0.018$ for Lasso yielding $5$ predictor variables $\{X_{2},X_{3}, X_{4}, X_{7},X_{8}\}$ with nonzero estimated coefficients, while the selected penalty parameter value is $0.222$ for RERF with $3$ predictors $\{X_{2},X_{3}, X_{4}\}$ used to generate the residuals in Step 2 of RERFs.

%

%
\section{Examples} \label{example}
\subsection{High-performance concrete strength example}\label{example1}
We use the high-performance concrete strength dataset  \cite{yeh1998modeling} as a real example to demonstrate our methodology. The concrete strength dataset is available on the UC Irvine Machine Learning Repository website, and has been widely used for evaluating machine learning algorithms, such as in \cite{xu2016case} and \cite{graczyk2010nonparametric}. It contains 1030 observations, with eight quantitative predictors (cement, water, fly ash, blast furnace slag, superplasticizer, coarse aggregate, fine aggregate and age of testing), and a response variable (concrete compressive strength). The Abrams rule \cite{popovics1990analysis} implies the approximate proportionality between the cement-to-water ratio (C/W) and the concrete compressive strength, so the cement-to-water ratio was computed as a predictor and included in the dataset.

Predicting concrete compressive strength (CCS) given amounts of ingredients and age is an important problem for civil engineering. However, high-performance concrete is such a highly complex material that modeling its strength behavior is very difficult. Exploratory analysis shows that the relationships between high-performance concrete strength and ingredients are nonlinear, and there are substantial interactions among predictors. Many previous studies have used machine learning algorithms, such as artificial neural networks and random forests, to tackle this problem. 

We compare the prediction performance of RERFs and RFs under six scenarios shown in Table \ref{concrete_scenario}. Scenario INT1 and INT2 are interpolation cases, while the rest are extrapolation cases. In Scenario INT1 and INT2, the complete dataset is randomly divided into training dataset and validation dataset. In Scenario EXT1 and EXT2, the complete dataset is divided based on the value of concrete compressive strength (CCS) so that the domains of CCS in the training dataset and validation dataset are disjoint. In Scenario EXT3 and EXT4, the cement-to-water ratios in the training dataset and validation dataset also have disjoint domains. We run 1000 simulations for each scenario. 


The computing results show that the average of selected penalty parameter values for Lasso is $0.08$, while the average of selected penalty parameter values for RERFs is $1.0$. Thus, there are fewer predictor variables that have nonzero estimated coefficients in Step 2 of RERFs than in Lasso.

\begin{figure}[ht]
	\centering
	\includegraphics[scale=0.7]{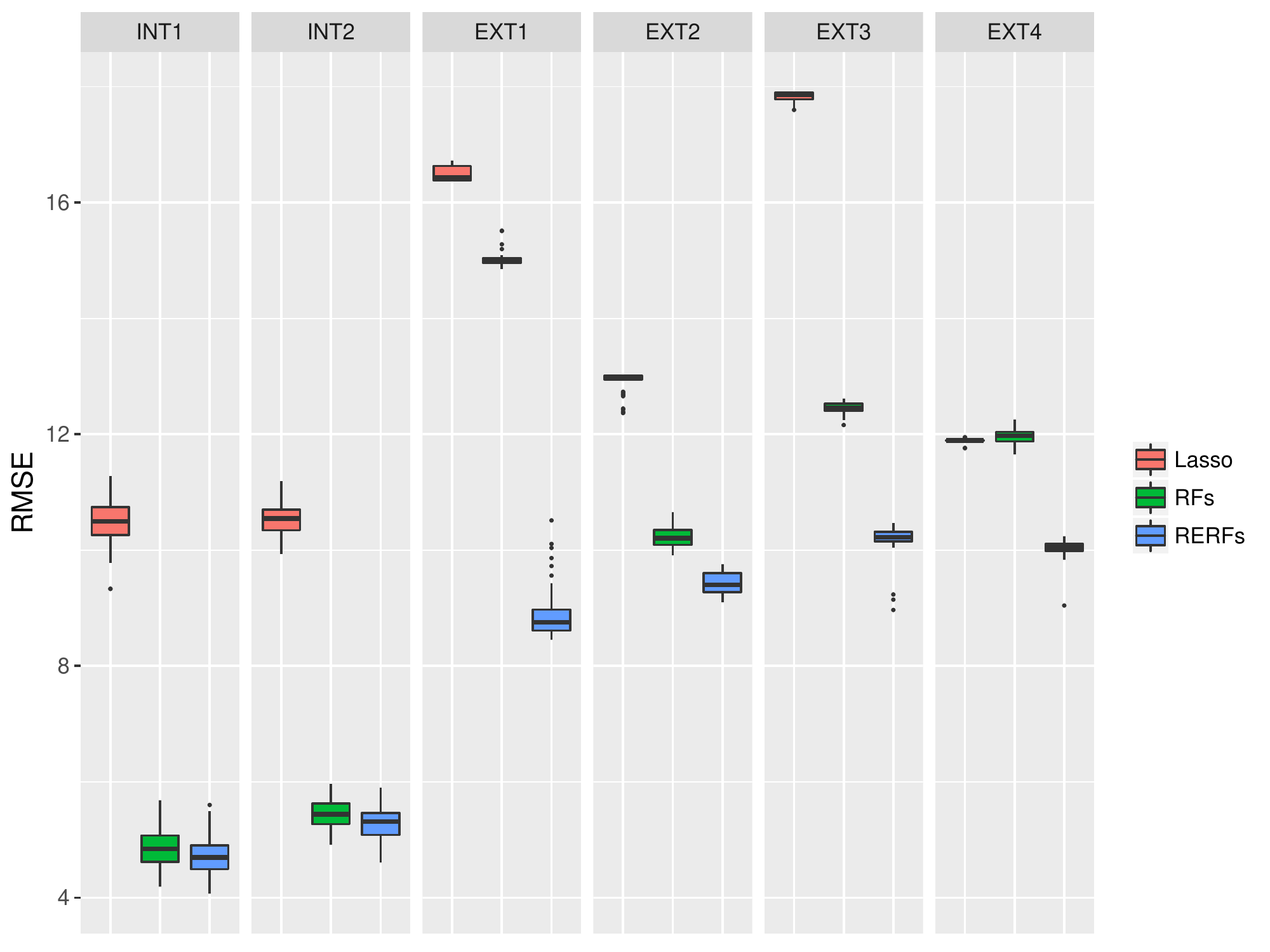}
	\caption{Boxplots of the RMSE values for Lasso, RFs and RERFs for each scenario in the concrete strength prediction example.}\label{concrete_rmse}
\end{figure}

Figure \ref{concrete_rmse} illustrates the RMSE values for Lasso, RFs and RERFs for each scenario, by which we can make the same conclusion as that in the simulation study. In the concrete strength prediction example, RERFs exhibits better prediction performance than RFs in both the interpolation and extrapolation cases, no matter whether Lasso is better than RFs or not. Particularly, in the extrapolation cases, RERFs approach far outperforms RFs. 

\begin{table}[ht]
	\centering
	\small
	\caption{Description of training and validation datasets in the concrete strength prediction example}\label{concrete_scenario}
	\begin{tabular}{lp{3.8cm}p{3.0cm}p{2.0cm}p{2.4cm}}
		\hline\hline
		Scenario & Training set& Validation set & \small{Training set sample size} & \small{Validation set sample size}\\ 
		\hline
		INT1& Random $3/4$ & Remaining $1/4$ & 772& 258\\
		INT2&Random $1/2$ & Remaining $1/2$ & 515& 515\\
		EXT1&CCS $>25$ &CCS $\leq 25$& 735& 295\\
		EXT2&CCS$<16$ or CCS$>56$ & $16\leq$ CCS $\leq 56$ &761& 269\\
		EXT3&C/W$<2$&C/W$\geq 2$ & 793&237\\
		EXT4&C/W$<1$ or C/W$>3$ & $1\leq$ C/W $\leq 3$ &804& 226\\
		\hline\hline
	\end{tabular}
\end{table}

\subsection{Iowa corn yield example}\label{example2}
Forecasting corn yield is an age-old and important problem in agriculture and economics. The United States produced roughly $14.2$ billion bushels of corn in the 2014-2015 crop marketing year, and the productions have been exported to more than 100 different countries. 
Iowa produces, by far, the most corn in the United States, supplying nearly 20 percent of the country's annual corn. Providing a valid corn yield prediction of Iowa before and within the harvesting season is of importance to agricultural policy decision, land planning, livestock husbandry, and option markets. 

In this section, we compare the performance of RERFs, RFs and Lasso in forecasting current year's corn yield in Iowa by using previous years' data and current year's meteorological and soil data. For instance, we used the complete data during 1988-2013 and the meteorological and soil data during January -- September 2014 to forecast the corn yield in 2014.

The dataset used in the analysis consists of three parts: corn yield data, meteorological data and soil data. The corn yield dataset is available on the National Agricultural Statistical Service (NASS) website. It contains annual averaged corn yields for grain per acre for 99 counties in Iowa from 1926 to 2015. The data are collected by annual survey after harvesting seasons. The meteorological dataset is available on \textit{Climate Data Online}  system of National Center for Environmental Information. It is station-based containing daily records of meteorological variables over most of meteorological monitoring stations in each county in Iowa from 1988 to 2015. The meteorological variables include temperature over surface, Drought Index, precipitation, air pressure, etc. We should note that the meteorological recordings are time series data. We regard the mean of the meteorological variable in each month as an individual predictor. For instance, we use the mean temperature for each month from May to September, so there are five monthly mean temperature predictors in the feature matrix. The soil dataset contains one predictor variable named \textit{Corn Suitability Rate}, a continuous variable measuring the corn productivity levels of soils, downloaded from the Iowa Soil Properties and Interpretations Database. There are $2689$ observations and $49$ predictor variables in the dataset. Observations with missing values were omitted in the analysis. 

Agricultural knowledge implies that extremely high and extremely low temperature may both cause low corn yield. Flooding (high precipitation) and drought (low precipitation) may cause low yield as well. As a consequence, we choose to include the quadratic terms of monthly mean temperature and Drought Index in the feature matrix $\bm{X}^{*}$ in Step 1 of RERFs. For all other predictors, only the first-order terms are included in $\bm{X}^{*}$. 

\begin{figure}[ht]
	\centering
	\includegraphics[scale=0.7]{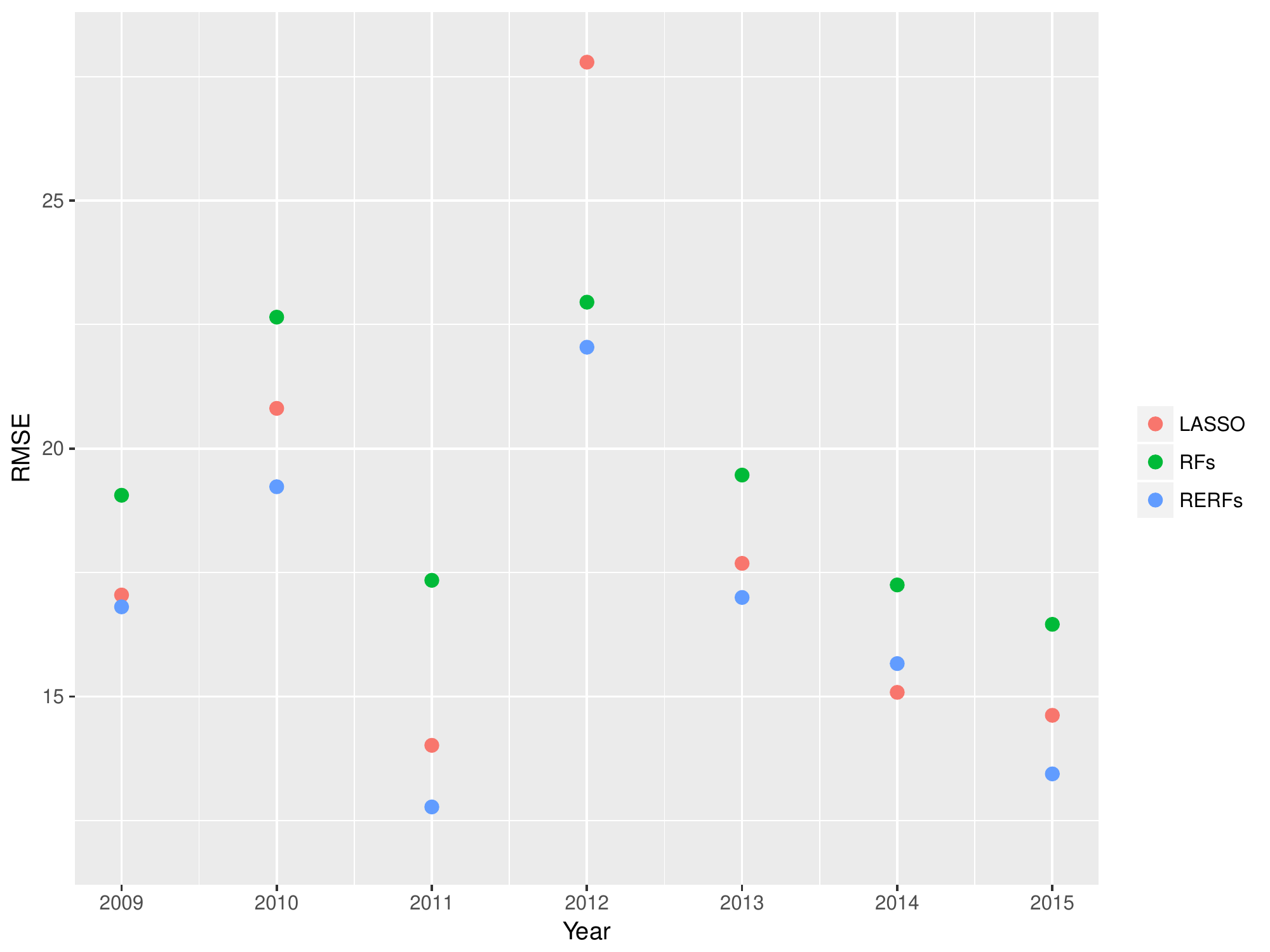}
	\caption{The RMSE values for Lasso, RFs and RERFs for each year from 2009 to 2015 in Iowa corn yield example.}\label{Iowa_rmse}
\end{figure}

The RMSE values for Lasso, RFs and RERFs are shown in Figure \ref{Iowa_rmse}. In the Iowa corn yield example, the prediction performance of RFs can be improved by using RERFs. The conclusion we make from Iowa corn yield example is consistent with the conclusion from simulation study and concrete strength prediction example. 

\section{Discussion}\label{discussion}
In this paper we have introduced the regression-enhanced random forest approach, a novel generalized RF method that has a better prediction performance than RFs in important situations often encountered in practice. The key idea of RERFs is borrowing the strength of penalized parametric regression to improve on nonparametric machine learning approaches. Specifically, for RERFs, we run Lasso before RF, then construct a RF on the residuals from Lasso. Because RERFs will be reduced to RFs for sufficiently large penalty parameter, RFs can be viewed as a special case of RERFs. 

Tuning parameter selection is critical to the performance of RERFs. The approach for selecting three tuning parameters have been discussed in detail in Section \ref{Method}. Because the exhaustive search on 3-dimensional tuning parameter space is time consuming, parallel computing can be applied to lessen the computing intensity. The other approach is an approximation procedure described as follows. First, fixing the nodesize and mtry to be the default values, we select a value of penalty parameter by cross validation. Second, using the selected value of penalty parameter $\lambda$, choose values of nodesize and mtry. Lastly, using the selected values of the nodesize and mtry obtained in the previous step, we update the value of the penalty parameter by cross validation. The above procedure reduces the computing intensity and yields values of the tuning parameter with cross-validation performance similar to parameters obtained by an exhaustive search. 

We focus on the comparison between RFs and RERFs in prediction performance for regression problems. As a fully nonparametric predictive algorithm, random forests may lack in efficiently incorporating known relationships between the response and the predictors. Moreover, random forests may fail in extrapolation problems where predictions are required at points out of the domain of the training dataset. However, RERFs can capitalize on the strength of both parametric and nonparametric methods and overcome the corresponding disadvantages. 

Numeric investigations, including one simulation study and two real data examples, all reach the same conclusion that the prediction performance of RERFs is better than that of RFs in both the interpolation and extrapolation problems we considered. Furthermore, in our extrapolation cases, RERFs far outperform RFs. Strategies analogous to those described here can be used to improve other machine learning methods via combination with penalized parametric regression techniques.



%



\vskip 0.2in
\bibliographystyle{plain}
\bibliography{RERFs}

\end{document}